\documentclass[conference]{IEEEtran}
\IEEEoverridecommandlockouts
\usepackage{cite}
\usepackage{amsmath,amssymb,amsfonts}
\usepackage{algorithmic}
\usepackage{graphicx}
\usepackage{textcomp}
\def\BibTeX{{\rm B\kern-.05em{\sc i\kern-.025em b}\kern-.08em
    T\kern-.1667em\lower.7ex\hbox{E}\kern-.125emX}}
 \usepackage{booktabs} 
\usepackage[bookmarks=false]{hyperref}
\usepackage{amsthm}
\usepackage{multirow}
\usepackage{footmisc}
\usepackage[linesnumbered,ruled,vlined]{algorithm2e}
\usepackage{caption}
\usepackage{subcaption}
\usepackage{mathrsfs}

\usepackage{stmaryrd}
\usepackage{diagbox}
\usepackage{xcolor}
\usepackage{balance}

\begin{document}
\newtheorem{theorem}{Theorem}

\title{Multi-Label Adversarial Perturbations}

\author{\IEEEauthorblockN{Qingquan Song, Haifeng Jin, Xiao Huang, Xia Hu} 
\IEEEauthorblockA{Department of Computer Science and Engineering, Texas A\&M University\\
\{song\_3134,jin,xhuang,xiahu\}@tamu.edu}
}

\maketitle

%
%

\begin{abstract}
Adversarial examples are delicately perturbed inputs, which aim to mislead machine learning models towards incorrect outputs. While most of the existing work focuses on generating adversarial perturbations in multi-class classification problems, many real-world applications fall into the multi-label setting in which one instance could be associated with more than one label. For example, a spammer may generate adversarial spams with malicious advertising while maintaining the other labels such as topic labels unchanged. To analyze the vulnerability and robustness of multi-label learning models, we investigate the generation of multi-label adversarial perturbations. This is a challenging task due to the uncertain number of positive labels associated with one instance, as well as the fact that multiple labels are usually not mutually exclusive with each other. To bridge this gap, in this paper, we propose a general attacking framework targeting on multi-label classification problem and conduct a premier analysis on the perturbations for deep neural networks. Leveraging the ranking relationships among labels, we further design a ranking-based framework to attack multi-label ranking algorithms. We specify the connection between the two proposed frameworks and separately design two specific methods grounded on each of them to generate targeted multi-label perturbations. Experiments on real-world multi-label image classification and ranking problems demonstrate the effectiveness of our proposed frameworks and provide insights of the vulnerability of multi-label deep learning models under diverse targeted attacking strategies.  Several interesting findings including an unpolished defensive strategy, which could potentially enhance the interpretability and robustness of multi-label deep learning models, are further presented and discussed at the end.
\end{abstract}


\section{Introduction}


Adversarial machine learning, which aims to enhance the security of machine learning models in adversarial settings, has attracted a lot of attentions in recent years. An intriguing property shared among vast models, especially for deep neural networks, is their vulnerabilities to adversarial examples, i.e., given malicious inputs with delicate perturbations, a model could be easily misled to generate erroneous decisions~\cite{szegedy2014intriguing,goodfellow2015explaining,liu2016delving}. The existence of adversarial examples could lead to severe impacts on the safe adoption of machine learning models in various applications, such as spammer filtering~\cite{wittel2004attacking,lowd2005good}, malware detection~\cite{grosse2016adversarial}, and autonomous driving~\cite{papernot2017practical}. It also motivates the explorations of generating powerful perturbations to analyze the vulnerability of machine learning models, detect their blind spots, and improve their robustness~\cite{barreno2010security,huang2011adversarial,papernot2016towards}. 

Existing work has been widely conducted in generating adversarial examples for multi-class classification algorithms~\cite{szegedy2014intriguing,goodfellow2015explaining, carlini2017towards, moosavi2016deepfool, papernot2016limitations}. Most of them are gradient-based approaches and have been proved to be effective on both targeted or non-targeted attacks. However, in many real-world applications, an instance may be associated with multiple labels. For example, in text categorization, a document may cover a range of topics, such as politics, economics, and diplomacy~\cite{hoi2006large}; in image classification, a neural scene image can contain both fields and mountains~\cite{boutell2004learning}. Different from the multi-class setting in which classes are mutually exclusive and one instance can only be assigned to one class (label), in multi-label learning, each instance can be associated with multiple labels,  thus opening more opportunities for attackers and leading to larger uncertainties for defenders.

Multi-label adversarial examples widely exist in both malicious and benign activities. From an attacker's perspective, he or she would like to alter certain labels while keeping some others unchanged for a more pertinent attack or camouflage. For instance, in an image recommendation system, an image spammer aims to generate spam images concerning specific topics such as sceneries targeting to certain users. In this case, the spammer expects the system classifier could misclassify the spam image as a benign object while correctly classifying it as a scenery image for the purpose of malicious recommendations~\cite{gupta2013faking}. From a benign perspective, the system administrators could leverage the perturbations to jointly perturb some sensitive attributes thus enhance the protection of users' privacy via preventing their personal information being inferred by malicious classifiers~\cite{jia2017attriinfer}. Moreover, multi-label adversarial examples could also be utilized for adversarial learning to improve the robustness of multi-label models~\cite{wu2017adversarial}.  Motivated by these observations, in this paper, we propose to investigate a novel and important problem, i.e., multi-label adversarial perturbation generation.




Generating targeted multi-label adversarial perturbations is still a rarely touched and challenging task. First, each instance might have an uncertain number of labels. It requires the attacking methods to be sensitive and discriminative in generating perturbations based on different targets. Second, coordinating multiple labels is difficult since labels are not mutually exclusive with each other. It is too arbitrary to target on a certain single label without considering the rests. Third, quantifying the attacking performance can be hard since multiple targeted labels may not be jointly achieved. These challenges prevent exiting attacking methods from being simply applied to generate multi-label adversarial perturbations. 

To tackle these challenges, in this paper, we propose a general attacking framework targeting on multi-label classification models, and conduct premier analysis on the perturbations for deep neural networks. Leveraging the ranking relationships among labels, we further introduce a ranking-based framework to attack multi-label ranking models. Through exploring the attacking performance targeting on manipulating different labels, we empirically validate the effectiveness of our frameworks and provide evidence for the vulnerability of multi-label deep learning models. Specifically, we aim to answer the following questions: (1) How to generate multi-label adversarial perturbations? (2) What is the performance of the generated perturbations based on different attacking strategies? The main contributions are summarized as follows: 

\begin{itemize}
\itemsep0.05em 
\item Formally define multi-label adversarial examples;
\item Propose a general framework and two corresponding methods to generate targeted multi-label adversarial perturbations for multi-label classification;
\item Propose a variation framework and two corresponding methods targeting on attacking multi-label ranking;
\item Examine the vulnerability of deep models in multi-label learning using various attacking methods and strategies.
\end{itemize}

\section{Preliminaries}
\noindent{\bf Notations}: 
Assume we have a multi-label classification problem with $l$ labels. Let $\mathbf{x} \in \mathbb{R}^{d\times 1}$ and $\mathbf{y} \in \{-1,1\}^{l\times1}$ denote the feature vector and label vector of an instance, respectively. In general, multi-label classification algorithms provide two types of outputs, either binary values that assign concrete relationships between instances and labels or confidence scores indicating the relevance of each instance with labels. We use function $H: \mathbb{R}^d \rightarrow \{-1,1\}^l$ to denote a multi-label \emph{\textbf{classifier}} which generates the first-type outputs, and $F: \mathbb{R}^d \rightarrow \mathbb{R}^l$ to denote a multi-label \emph{\textbf{predictor}} which predicts continuous relevance scores corresponding to the second-type outputs. Both $H$ and $F$ can be decomposed as $n$ sub-functions, i.e., $H=\{ h_1, \ldots, h_l \}$ and $F=\{ f_1, \ldots, f_l \}$, where $h_j(\mathbf{x})$ and $f_j(\mathbf{x})$ indicate their predictions of $y_{j}$. Based on certain classification thresholds, it is easy to induce $H$ from $F$, e.g., $h_j(\mathbf{x}) = \mathbf{I}_{ \llbracket f_j(\mathbf{x}) > t_j(\mathbf{x}) \rrbracket }-\mathbf{I}_{ \llbracket f_j(\mathbf{x}) < t_j(\mathbf{x})}$, where $\mathbf{I}_{ \llbracket \cdot \rrbracket }$ is an indicator function. Following the notations, we formally define two categories of multi-label adversarial examples.

Given an instance $\textbf{x}$, let $H$ be a classifier satisfying $H(\mathbf{x})=\textbf{y}$, where $\textbf{y}$ is the ground truth labels of $\textbf{x}$. Here, we assume that $H$ can correctly classify all labels of $\mathbf{x}$. Though it may not be true in practice, one could simply achieve this by limiting $L$ to cover only the labels that $H$ correctly classifies $\textbf{x}$ into. Then we have the following two cases:

\noindent\textbf{A Non-targeted Multi-label Adversarial Example} of $H$ around $\textbf{x}$ is defined as an instance $\textbf{x}^\ast$ that satisfies:
\begin{itemize}
\itemsep-0.1em 
\item[1.] $\textbf{x}^\ast$ is close to $\textbf{x}$ under a certain distance measure;
\item[2.] $\textbf{x}^\ast$ has the same ground truth labels with $\textbf{x}$;
\item[3.] $H(\textbf{x}^\ast) \neq  \textbf{y}$, i.e., $\exists i \in L$ such that $h_i(\mathbf{x}^\ast) \neq y_i$.
\end{itemize}

\noindent\textbf{A Targeted Multi-label Adversarial Example} of $H$ around $\textbf{x}$ is defined as an instance $\textbf{x}^\ast$ that satisfies:
\begin{itemize}
\itemsep-0.1em 
\item[1.]  $\textbf{x}^\ast$ is close to $\textbf{x}$ under a certain distance measure;
\item[2.] $\textbf{x}^\ast$ has the same ground truth labels with $\textbf{x}$;
\item[3.] Given label sets A and B, $h_a (\textbf{x}^\ast) \neq y_a, \forall a\in A$; $h_b (\textbf{x}^\ast) = y_b, \forall b\in B$. ($A\neq \emptyset$, $A \cup B\subset L$, $A \cap B = \emptyset$)
\end{itemize}

The first two characteristics for both categories of examples are motivated by the existing work with multi-class settings~\cite{szegedy2014intriguing,goodfellow2015explaining,liu2016delving,carlini2017towards}. Both of them indicate that the difference between an adversarial example and the original example should be indistinguishable for humans, i.e., the perturbation should be small. From the last characteristics, we can see that for non-targeted adversarial examples, there is no specific label that we target to manipulate which is similar to the non-targeted attack in multi-class cases; but for targeted examples, two label sets $A$ and $B$ are specified corresponding to the labels we intend to vary and the labels we expect to fix. This definition is motivated by the observations that, in practice, attackers are often only interested in attacking a specific set of labels $(i.e., A)$ while keeping another specific set of labels $(i.e., B)$ for a more pertinent attack or camouflage. Moreover, as each label is binary, there is no need to specify the values we aim to achieve for each $h_a (\textbf{x}^\ast), a \in A$. In the rest of the paper, we will focus on investigating the targeted multi-label adversarial examples, since they are more practical and meaningful in real-world systems. For the ease of presentation, we use ``adversarial example generation'' and ``adversarial perturbation generation'' interchangeably.


\section{Targeted Multi-label Adversarial Perturbation Generation}
To cope with general multi-label settings while conducting effective attacks, we propose two joint attacking frameworks towards attacking multi-label learning models with classification and ranking purposes, respectively. We first introduce a general classification-targeted framework, which aims to manipulate a specific set of predicted labels in the multi-label classification problem. The key idea is to construct a new classification problem by reversing the relationship between instances and classifiers, and generate perturbations through optimization. Motivated by the broad applications of multi-label ranking techniques, we then tailor the framework towards a ranking-targeted framework to attack models with ranking purpose. The core idea is to rerank the predicted scores based on the targeted labels to construct a new multi-label ranking problem. For both frameworks, we separately design two specific methods in subsequent sections.

\subsection{The General Frameworks of the Classification-targeted and Ranking-targeted Attacks}
\label{sec:general}
In this section, we introduce the two types of frameworks we proposed for attacking multi-label classification models and multi-label ranking models, respectively.

\subsubsection{Type I. Classification-targeted Framework}

We first investigate generating adversarial examples for multi-label classification models. Since each label is either $-1$ or $1$, for a targeted classifier $H$ and a benign instance $\textbf{x}$, our goal can be mathematically formulated as follows,
\begin{equation}
\begin{aligned}
\centering
& \underset{ \textbf{r} }{\text{minimize}}  && \|\mathbf{r}\| \\
& \text{subject to} && h_{a}(\mathbf{x}+\mathbf{r}) = -y_{a}, \, a\in A, \\
& && h_{b}(\mathbf{x}+\mathbf{r}) = y_{b}, \, {b}\in B,
\end{aligned}
\label{equ:type1}
\end{equation}
where $\textbf{r}$ is the expected perturbation. $\textbf{x}^{\ast} = \textbf{x}+\textbf{r}$ is the generated adversarial example. $\| \cdot \|$ denotes a certain norm which is usually defined as the $L_p$ norm, where $p$ is chosen based on specific settings or demands~\cite{carlini2017towards}. %

\subsubsection{Type II. Ranking-targeted Framework}
\label{sec:rank_general}

\renewcommand{\arraystretch}{1.4}
 \begin{table}[t!]\scriptsize
 \vspace{5pt}
\centering
\setlength{\tabcolsep}{7pt}
\caption{Two divisions of labels of the target sample $\textbf{x}$.}
  \begin{tabular}{|c|c|c|c|}
    \hline
      \diagbox{Ground Truth \\ Division}{Attack Division} & $A$ & $B$   & $C=L\setminus A \cup B$  \\
      \hline
    $Y_1=\{y_i| y_i=1, i\in L\}$  & ${\color{blue} A_1}$ & ${\color{red} B_1}$ &  $C_1$ \\
    \hline
     $Y_{-1}=\{y_i| y_i=-1, i\in L\}$ & ${\color{red} A_{-1}}$ & ${\color{blue} B_{-1}}$ &  $C_{-1}$ \\
 \hline
\end{tabular}
\vspace{-5pt}
\label{table:div}
  \end{table}
Existing work on the ranking-based prediction techniques has demonstrated their effectiveness in both multi-label classification and ranking tasks~\cite{tang2009large,agrawal2013multi}. It motivates us to explore the corresponding adversarial examples to evaluate the robustness of these models. A straightforward way to attack them is to generate examples based on our first type of framework. However, two problems may perplex it. (1) Soft thresholds: since ranking-based models aim at producing ranking relationships among labels, no hard thresholds need to be specified. (2) Label relationships: the classification-targeted framework does not explicitly take the relationships among labels into account which may not be generalizable for attacking ranking-based algorithms. To tackle the issues, we propose a ranking-targeted framework. Mathematically, given a predictor $F$, we target at, 
\begin{equation}
\begin{aligned}
\centering
& \underset{ \textbf{r} }{\text{minimize}}  && \|\textbf{r}\| \\
& \text{subject to} && f_\alpha (\textbf{x}+\textbf{r}) \leq f_\gamma (\textbf{x}+\textbf{r}) \leq f_\beta (\textbf{x}+\textbf{r}), \\
&&& \forall \alpha \in A_1 \cup B_{-1}, \, \beta \in A_{-1} \cup B_1, \, \gamma \in C.
\end{aligned}
\label{equ:general2}
\end{equation}

These constraints are motivated by two divisions of the whole label set $L$ based on the target instance $\textbf{x}$ and the attacking strategy. As shown in Table~\ref{table:div}, the header sets of the row and column represents the two types of divisions, i.e., the first column denotes a division using the ground truth labels of instance $\textbf{x}$ and the first row denotes a division using the attacking sets $A$ and $B$. Each grid in the table denotes the intersection of corresponding header sets of its row and column, e.g., $A_1=A\cap Y_1$. Labels in the set $A_{-1} \cup B_1$ should be positive after attacking, and labels in the set $A_1 \cup B_{-1}$ should be negative. The predictions of the unconcerned labels $\gamma\in C$ are put in the middle to reinforce the attacking ability. It could accentuate the ranking gap between labels on the two sides of the inequality. This constraint may become pretty harsh especially when $|A|+|B| \ll |C|$, where $|\cdot|$ represents the number of elements in a given set. To obtain relatively mild constraints, we can limit $\gamma$ within other label sets such as $C_{-1}$, $C_1$, or $\emptyset$, or directly fix a hard threshold. 

\subsubsection{Connection of the Two Frameworks}
The proposed two frameworks are highly connected with each other. Comparing Equation~\eqref{equ:type1} and~\eqref{equ:general2}, if we replace the soft thresholds $f_\gamma (\textbf{x}+\textbf{r})$, with hard classification thresholds, then these two types of frameworks are equivalent to each other. This means Type II framework could be more general if we do not know the specific classification thresholds under partially black-box settings. By contrast, the Type I framework directly takes the classification thresholds into account. Thus, if we conduct a pure white-box attack on multi-label classification algorithms and do not use hard thresholds in Type II framework, the Type I framework may achieve better performance since it takes more accurate classification thresholds into account. Based on these two types of frameworks, now we propose several attacking methods.

\subsection{ Type I. Attack Multi-label Classification}
In this section, we propose two attacking methods based on the Type I framework. To avoid the intricacy induced by classification thresholds, we fix certain thresholds and express the constraints using predictor $F$. For example, for a linear predictor $F(\mathbf{x})$, the corresponding classifier can be induced with threshold $0$ as $H(\mathbf{x})=sgn(F(\mathbf{x}))$, where $sgn(\cdot)$ is the signum function. So the constraints in Equation~\eqref{equ:type1} are equivalent to Equation~\eqref{eq:linear}. For a neural network with sigmoid output layer and $0.5$ threshold (i.e., $H(\mathbf{x})=sgn(F(\mathbf{x})-0.5)$), the equivalent constraints are described in Equation~\eqref{eq:nn}.
\begin{subequations}
    \begin{align}
       & y_a f_i(x+r)\le 0,          && -y_b f_b(x+r)\le 0;   \label{eq:linear}   \\
       & y_a f_i(x+r)\le 0.5y_a, &&  -y_b f_b(x+r)\le -0.5y_b.   \label{eq:nn}
    \end{align}
\end{subequations}

\noindent Therefore, Equation~\eqref{equ:type1} could be transformed as follows:
\begin{equation}
\begin{aligned}
\centering
& \underset{ \textbf{r} }{\text{minimize}}  && \|\mathbf{r}\| \\
& \text{subject to} && \mathbf{y}'\odot F'(\mathbf{x}+\mathbf{r}) \le \mathbf{c}, \\
\end{aligned}
\label{equ:type1_2}
\end{equation}
where $\mathbf{y'}=[y_{a_1},\ldots,y_{a_{|A|}}, -y_{b_1},\ldots,-y_{b_{|B|}}]^\top$ and $F'=[f_{a_1},$$\,\ldots,f_{a_{|A|}}, f_{b_1},\ldots,f_{b_{|B|}} ]^\top$. $\odot$ is the Hadamard product. Vector $\mathbf{c}\in \mathbb{R}^{|A|+|B|}$ is defined based on the thresholds, targeted labels, and the predictor model we attack, e.g., $\mathbf{c}=\mathbf{0}$ represents linear predictors with threshold $0$. 

\subsubsection{Multi-label Carlini \& Wagner Attack (ML-CW)} A straightforward way to solve Equation~\eqref{equ:type1_2} is to convert the constraints to regularizers such as using the hinge loss:
\begin{equation}\small
\begin{aligned}
\centering
& \underset{ \textbf{r} }{\text{minimize}}  && \|\mathbf{r}\| +\lambda \sum_{i=1}^{|A|+|B|}  \max(0, \mathbf{y}'_iF_i'(x+r)-c_i ),
\end{aligned}
\label{equ:type1_3}
\end{equation}
where $\lambda$ is a trade-off penalty between the perturbation size and attacking accuracy. The hinge loss is chosen here to provide a soft margin for each constraint, which has been proved to be effective in multi-class attacking~\cite{carlini2017towards}.  Since we target on the white-box attack here, the optimization can be done based on the algorithms we target on attacking, e.g., gradient decent methods for neural networks. To alleviate the influence of the hyperparameter selection, we employ the binary search approach~\cite{carlini2017towards} to select $\lambda$. In each binary search iteration, $\lambda$ is either magnified decuple or shrunk by half depending on whether the attack is successful or not. We choose the best perturbation, which satisfies the largest number of constraints in Equation~\eqref{equ:type1_2}. If two perturbations satisfy the same number of constraints, the one that has a smaller distortion is better. 

\subsubsection{Multi-label DeepFool Attack (ML-DP)} 
\setlength{\textfloatsep}{15pt}
\begin{algorithm}[t!]
\DontPrintSemicolon
\KwIn{$\mathbf{x}$, predictor $F$, classifier $H$, target label sets $A$ and $B$, maxiter.} 
 \KwOut{ $\mathbf{r}^\ast$, $\mathbf{x}^\ast $.}
Initialize the best adversarial example as: $\mathbf{x}^{\ast}=\mathbf{x}$; \;
Initialize $\mathbf{x}_0=\mathbf{x}$, $\mathbf{r}_{0}=\mathbf{0}$, and $i=0$;\;
Define $F'$, $\mathbf{y}'$ and $\mathbf{c}$ based on the inputs;\;
 \While { $i \leq \text{maxiter} $}{
Calculate $\mathbf{P}(\mathbf{x}_i)$ and $\mathbf{q}(\mathbf{x}_i)$ based on Equation~\eqref{equ:fool2};\;
$\Delta\mathbf{r}  = \mathbf{P}(\mathbf{x}_i)(\mathbf{P}(\mathbf{x}_i)^\top\mathbf{P}(\mathbf{x}_i) )^{-1} \mathbf{q}(\mathbf{x}_i) $;\;
// Update current adversarial example \& perturbation: \;
$\mathbf{x}_{i+1}=\mathbf{x}_i+\mathbf{r}_i$,$\quad\mathbf{r}_{i+1}=\mathbf{r}_i+\Delta\mathbf{r}$;\;
// Find the sets of constraints that $\mathbf{x}_{i+1}$ and $\mathbf{x}^{\ast}$ satisfy in Equation~\eqref{equ:type1_2}:\;
$Cst_i = \{k|  \mathbf{y}'_k\circledast F_k'(\mathbf{x}_{i+1}) \le \mathbf{c}_k, k\in 1,\ldots, |A|+|B|  \}$;\;
$Cst_{b} = \{k|  \mathbf{y}'_k\circledast F_k'(\mathbf{x}^{\ast}) \le \mathbf{c}_k, k\in 1,\ldots, |A|+|B| \}$;\;
\If {  $|Cst_i| > |Cst_{b}|$ }{
$\mathbf{x}^{\ast}= \mathbf{x}_{i}$, $\mathbf{r}^\ast = \mathbf{r}_{i}$;
}\ElseIf {$|Cst_i| = |Cst_{b}|$ \& $ \|\mathbf{x}_{i}\|_2 < \|\mathbf{x}^{\ast} \|_2$ }{
$\mathbf{x}^{\ast}= \mathbf{x}_{i}$, $\mathbf{r}^\ast = \mathbf{r}_{i}$;
}
$i=i+1$;\;
}
\Return  $\mathbf{r}^\ast$, $\mathbf{x}^{\ast}$.
\caption{Multi-label DeepFool Attack (ML-DP)}
\label{alg:fool}
\end{algorithm}

Considering the high nonlinearity of the constraints, a variation of solving Equation~\eqref{equ:type1_2} is to utilize the linear approximation of $F'$ to linearized the constraints as:
\begin{equation}
\begin{aligned}
\centering
& \underset{ \textbf{r} }{\text{minimize}}  && \|\mathbf{r}\| \\
& \text{subject to} && \mathbf{y}'\odot [ F'(\mathbf{x}) + (\frac{ \partial F'}{ \partial \mathbf{x}})^\top \mathbf{r} ] \le \mathbf{c},
\end{aligned}
\label{equ:fool}
\end{equation}
where $\frac{ \partial F'}{ \partial x}= [\frac{\partial f_{a_1}}{\partial \mathbf{x}}, \ldots, \frac{\partial f_{a_{|A|} }}{\partial \mathbf{x}}, \frac{\partial f_{b_1}}{\partial \mathbf{x}}, \ldots, \frac{\partial f_{b_{|B|} }}{\partial \mathbf{x}}] \in \mathbb{R}^{d \times (|A|+|B|)}$. Similar linearization of constraints have been proved to be effective in the multi-class attacking~\cite{moosavi2016deepfool}. However, generalizing them to the multi-label attack is not straightforward since multiple labels bring more complex restrictions to the perturbation $\mathbf{r}$. We provides a simple approach here by greedily solving a set of underdetermined linear equations. It is easy to see that the above problem is equivalent to minimizing $\mathbf{r}$ under a set of linear constraints with varying coefficients:
\begin{equation}
\begin{aligned}
\centering
& \underset{ \textbf{r} }{\text{minimize}}  && \|\mathbf{r}\|_2^2 \\
& \text{subject to} && \mathbf{P}(\mathbf{x})^{\top}\mathbf{r} \le \mathbf{q}(\mathbf{x}),
\end{aligned}
\label{equ:fool2}
\end{equation}
where $\mathbf{P}(\mathbf{x})=( \mathbf{1}_d \times  \mathbf{y'}^\top )\odot \frac{\partial F'(\mathbf{x})}{\partial \mathbf{x}}$ and $\mathbf{q}(\mathbf{x})=\mathbf{c}-\mathbf{y'}\odot F'(\mathbf{x})$. Here we measure the perturbation with $L_2$ norm for its simplicity and generality. The optimization is quite difficult because $\mathbf{P}(\mathbf{x})$ and $\mathbf{q}(\mathbf{x})$ vary with the changing of the perturbation. As feature dimension $d$ is usually larger than the number of labels $l$, i.e., $d\gg |A|+|B|$, the system is underdetermined for any fixed coefficients. Since in each iteration, the coefficients are uncorrelated with the current sub-perturbation $\mathbf{r}$, we propose to solve it in a greedy manner using a pseudo-inverse of $\mathbf{P}(\mathbf{x})$. The whole algorithm is shown in Algorithm~\ref{alg:fool}. We first define $F'(\mathbf{x})$, $\mathbf{y}'$, and $\mathbf{c}$ based on the inputs, and then greedily solve the underdetermined linear system with varying coefficients. After updating the current adversarial example $\mathbf{x}_{i+1}$ and the perturbation $\mathbf{r}_{i+1}$ in each iteration, we decide whether accepting $\mathbf{x}_{i+1}$ and $\mathbf{r}_{i+1}$ or not based on the constraints in Equation~\eqref{equ:type1_2} that $\mathbf{x}_{i+1}$ and $\mathbf{x}^\ast$ satisfy, i.e., $Cst_i$ and $Cst_b$.  It should be noted that this greedy algorithm does not guarantee a convergence to the optimal perturbation and may even not converge. Thus, we fix the maximum number of iterations and select the optimal perturbations by jointly consider the number of constraints they satisfied and the size of their distortions.


\subsection{Type II. Attack Multi-label Ranking}
In this section, we propose two more methods leveraging the relationships among labels based on the Type II framework. This type of attack is a more general ranking-based attack leveraging the relationships among labels.  


\subsubsection{Rank I Attack}  Assume $\Omega^{-}=A_1\cup B_{-1}$, $\Omega^{+}=A_{-1}\cup B_1$, and $\Omega^\circ=C$. Motivated by the empirical multi-label ranking loss in~\cite{schapire2000boostexter}, we proposed to minimize the average fraction of misordered label pairs. By adopting hinge loss similar to ML-CW, an alternative loss function can be formulated as:
\begin{equation}\scriptsize
\begin{aligned}
\centering
 \mathcal{L}_0 =  \|\textbf{r}\| 
&+  \frac{1}{ |\Omega^{-}||\Omega^{+}|  } \sum_{(\alpha,\beta) \in \Omega^{-} \times \Omega^{+}}      \max(0,   e^{ f_{\alpha} (\mathbf{x}+\mathbf{r})} - e^{f_{\beta} (\mathbf{x}+\mathbf{r})} ) \\
&+ \frac{1}{ |\Omega^{-}||\Omega^{\circ}|  } \sum_{(\alpha,\gamma) \in \Omega^{-} \times \Omega^\circ}  \max(0,  e^{ f_{\alpha} (\mathbf{x}+\mathbf{r})}- e^{f_{\gamma} (\mathbf{x}+ \mathbf{r}) } ) \\
&+  \frac{1}{ |\Omega^{\circ}||\Omega^{+}|  }  \sum_{(\gamma, \beta) \in \Omega^{\circ} \times \Omega^{+}}   \max(0, e^{ f_{\gamma} (\mathbf{x}+\mathbf{r}) } - e^{ f_{\beta} (\mathbf{x}+\mathbf{r}) }  ).
\end{aligned}
\label{equ:rank1}
\end{equation}
The exponential function is chosen to severely penalize the ranking errors inspired by~\cite{zhang2006multilabel}. When the number of labels is large, computing this equation would be time-consuming.
To reduce the time complexity without losing much attack power, Equation~\eqref{equ:rank1} could be simplified by extracting the maximum and minimum predictions of labels in each label set as follows:
\begin{equation}\small
\begin{aligned}
\centering
\mathcal{L}_1 = \|\textbf{r}\| 
& +  \lambda_1\max(0, \max_{\alpha\in \Omega^{-}}e^{f_{\alpha}(\mathbf{x}+ \mathbf{r} )} - \min_{\beta\in \Omega^{+}}e^{f_{\beta}(\mathbf{x}+ \mathbf{r} )}  ) \\
& + \lambda_2\max(0, \max_{\alpha\in \Omega^{-}}  e^{f_{\alpha}(\mathbf{x}+ \mathbf{r} )} - \min_{\gamma\in \Omega^{\circ}}e^{f_{\gamma}(\mathbf{x}+ \mathbf{r} )} ) \\
& +\lambda_3 \max(0, \max_{\gamma\in \Omega^\circ}e^{f_{\gamma}(\mathbf{x}+ \mathbf{r} )} - \min_{\beta\in \Omega^{+}}e^{f_{\beta}(\mathbf{x}+ \mathbf{r} )} ),
\end{aligned}
\label{equ:rank2}
\end{equation}
where $\lambda_i (i=1,2,3)$ are hyperparameters that make a trade-off among terms. This loss function could be further simplified based on the choice of $\Omega^\circ$. If $\Omega^\circ \neq \emptyset$, the first term could be reduced; otherwise the last two terms are nonexistent.

\subsubsection{Rank II Attack}  
\label{sec:rank2}
Though to some extent, Rank I attack takes the label correlation into account, it may not be sensitive enough in certain cases. For example, if we have a benign instance $\mathbf{x}$ with ground truth labels $[-1,1,1]^\top$ and targeted attacking labels $[-1,-1,1]^\top$, the label probabilities predicted by the predictor $F$ could be $[0.01,0.98,0.99]^\top$. In this case, Rank \uppercase\expandafter{\romannumeral1} cannot provide successful attack since Equation~\eqref{equ:rank2} is equal to $0$ for the benign instance $\mathbf{x}$ and no loss is suffered. This problem comes from the rank-constraint of the general framework defined in Equation~\eqref{equ:general2}. Based on this constraint, no discrimination is made for labels in set $\Omega^{-}=A_1\cup B_{-1}$. Similar situation happens for $\Omega^{+}= A_{-1} \cup B_1$. To solve this problem, we add two more constraints defined as follows:
\begin{equation}
\begin{aligned}
\centering
 \quad & f_{\alpha_1} (\textbf{x}+\textbf{r}) \leq f_{\alpha_2} (\textbf{x}+\textbf{r}),\quad \forall \alpha_1 \in A_1,\, \alpha_2\in B_{-1}, \\
 \quad & f_{\beta_1} (\textbf{x}+\textbf{r}) \leq f_{\beta_2} (\textbf{x}+\textbf{r}),\quad \forall \beta_1 \in B_{1},\, \beta_2 \in A_{-1}.
\end{aligned}
\label{equ:rank_constraint}
\end{equation}
The first constraint forces the probabilities of newly added negative labels to be smaller than all negative labels, while the second highlights the new positive labels. Leveraging the two constraints, the loss of Rank \uppercase\expandafter{\romannumeral2} attack is defined as:
\begin{equation}\small
\begin{aligned}
\centering
\mathcal{L}_2 = \mathcal{L}_1
& +  \lambda_4\max(0, \max_{\alpha_1\in A_{1} } e^{f_{\alpha}(\mathbf{x}+ \mathbf{r} )} - \min_{\alpha_2\in B_{-1}} e^{f_{\beta}(\mathbf{x}+ \mathbf{r} )} ) \\
& + \lambda_5\max(0, \max_{\beta_1\in B_{1} }e^{f_{\beta}(\mathbf{x}+ \mathbf{r} )} - \min_{\beta_2\in A_{-1}} e^{f_{\beta}(\mathbf{x}+ \mathbf{r} )} ),
\end{aligned}
\label{equ:rank4}
\end{equation}
where $\mathcal{L}_1$ is the loss function of Rank \uppercase\expandafter{\romannumeral1} attack. Similar to ML-CW, we use binary search for both Rank I and Rank II to determine a suitable $\lambda_i$. All the $\lambda_i$ are defined as the same in the experiments for simplicity. To select the best perturbation, after finding the optimal perturbation $\mathbf{r}_{\lambda}$ and its corresponding adversarial example $\mathbf{x}_{\lambda}^\ast$ for each $\lambda$,  we utilize Kendall $\tau_b$ ranking correlation coefficient~\cite{agresti2011categorical} to quantify the similarity between the new label prediction $\mathbf{y}_{\lambda}^\ast$  and a criterion vector $\mathbf{y}^{\tau} \in \{-2,-1,0, 1, 2\}^{l\times1}$. $\mathbf{y}^{\tau}$ is defined based on the rank constraints in Equations~\eqref{equ:general2} and~\eqref{equ:rank_constraint}, i.e., for any $i=1,2,\ldots,l$, $\mathbf{y}^{\tau}_i=-2,-1,0,1,2$ correspond to $i \in A_{1},B_{-1},C, B_{1}, A_{-1}$, respectively. The higher the similarity is, the better the perturbation fits the constraints. If two perturbations have the same $\tau_b$ score, the smaller one would be selected.

\section{Experiments}
We empirically evaluate the performance of the proposed attacking methods and explore the generated perturbations targeting on different labels. Three major aspects are analyzed:

\noindent \textbf{Q1}: What is the general performance of different methods? 

\noindent \textbf{Q2}: What is the performance of different attacking methods with specified targeted labels?

\noindent \textbf{Q3}: What are the characteristics of the generated multi-label adversarial perturbations?

%
%






%
%
\renewcommand{\arraystretch}{1.2}
 \begin{table}[t!]\small
\centering
\setlength{\tabcolsep}{3.7pt}
\vspace{3pt}
 \caption{Dataset statistics.}
\begin{tabular}[0.1\textwidth]{cccc}\toprule
& \# of Labels & Retraining $+$ Validation  & Testing \\ 
    \hline
     VOC 2007 & 20    & 5011  &  4952  \\
    VOC 2012 & 20  &   5717 &  5823 \\
\bottomrule
\end{tabular}
\vspace{2pt}
\label{Datasets}
  \end{table}

\subsection{Datasets}

The experiments are conducted on two different multi-label image datasets, i.e., PASCAL VOC 2007~\cite{everingham2010pascal} and VOC 2012~\cite{everingham2015pascal}.  Both of them are benchmark datasets, which are widely adopted in various multi-label classification/ranking work~\cite{perronnin2010improving,Chatfield14,wang2016cnn}. Table~\ref{Datasets} shows the basic statistics two datasets. For VOC 2012, we split the original training set into the retraining and validation sets, and use the original validation set as the testing set since the ground truth labels of the original testing images are not officially provided.

\subsection{Targeted Multi-label Learning Model} 


We focus on attacking deep neural networks given its superior performance in multi-label learning~\cite{wang2016cnn,wei2016hcp,shen2017learning} and vulnerability to adversarial examples~\cite{szegedy2014intriguing}. The classifiers (and predictors) are built upon the Inception v3 network~\cite{szegedy2016rethinking} pre-trained on ImageNet dataset~\cite{russakovsky2015imagenet}. Since inception v3 is constructed for multi-class classification, it cannot be directly applied to multi-label cases. Thus, we use a similar idea described in~\cite{Chatfield14} to retrain the model by replacing softmax layers with sigmoid classification layers.  To ensure prediction performance, both the instance-wise and label-wise losses are considered~\cite{wu2016unified}. 
We combine the instance-wise ranking loss motivated by the widely used multi-label neural network BPMLL~\cite{zhang2006multilabel}, and the label-wise AUC score~\cite{wu2016unified} to build up the loss function as follows:
\begin{equation}\small
\begin{aligned}
\centering
& \mathcal{J} = && \lambda \frac{1}{n}\sum_{i=1}^n \frac{1}{|Y_{i\cdot}^{+}| |Y_{i\cdot}^{-}|}  \sum_{(p,q) \in Y_{i\cdot}^{+} \times Y_{i\cdot}^{-} } \exp(f_q(\mathbf{x}_i)-f_p(\mathbf{x}_i)) \\
& && + \frac{1}{l} \sum_{j=1}^l \frac{1}{|Y_{\cdot j}^{+}| |Y_{\cdot j}^{-}|  }  \sum_{(p,q) \in Y_{\cdot j}^{+} \times Y_{\cdot j}^{-} } \exp(f_j(\mathbf{x}_q)-f_j(\mathbf{x}_p)). 
\end{aligned}
\label{equ:targeted_model}
\end{equation}
The first term is the modified instance-AUC score used in BPMLL, where $Y_{i \cdot}^{+} = \{j| y_{ij}=1\}$ and $Y_{i \cdot}^{-} = \{j| y_{ij}=-1\}$. This term is used to extract label relationships for each instance. The second term is the label-wise ranking score, where $Y_{ \cdot j}^{+} = \{i| y_{ij}=1\}$ and $Y_{ \cdot j}^{-} = \{i| y_{ij}=-1\}$. This term is used to alleviate the label-imbalance problem existed in both of the datasets. The trade-off parameter $\lambda$ is chosen to be $0.5$ in our experiments based on the validation dataset. We select the batch-size to be 100. After the retraining process, the final testing performance regarding five commonly used measures~\cite{wu2016unified} is reported in Table~\ref{OriginalAcc}. Partial retraining details are referred to~\cite{code}. It is worth noted that we do not focus on attacking some conventional multi-label classification models such as Binary Relevance~\cite{tsoumakas2006multi} here due to the fact that: they are both label transformation based method, which could be separated as several uncorrelated multi-class classification components and easily attacked by traditional multi-class attacking methods. Also, as this is a very early work on attacking multi-label learning algorithms, we intend to use the state-of-the-art and most representative methods and leave the rest for future exploration.

\renewcommand{\arraystretch}{1.2}
 \begin{table}[t!]\small
 \vspace{3pt}
\centering
\setlength{\tabcolsep}{3.7pt}
 \caption{Original accuracy of model to be attacked.} 
\begin{tabular}[0.1\textwidth]{ccccc}\toprule
   {} & Hamming  & macro/micro-F1 & Ranking Loss & mAP \\ 
    \hline
     VOC 2007 &0.0504 &  0.7278/0.7182  &  0.0175  &  0.9239 \\ 
        VOC 2012 &0.0491 &  0.7340/0.7252  &  0.0166 & 0.9320  \\
\bottomrule
\end{tabular}
\vspace{2pt}
\label{OriginalAcc}
  \end{table}

\subsection{Attacking Methods}

Besides the four proposed iterative methods, to give a more comprehensive analysis of different types of multi-label attacking methods, we extend two widely adopted multi-class attacking methods into multi-label settings and summarize all six methods analyzed in our experiments as follows:

\begin{itemize}
\item \textbf{Targeted Fast Gradient Sign Method (FGS)}~\cite{goodfellow2015explaining}\textbf{:} 
We extend this multi-class attacking method to multi-label setting via the following equation:
\begin{equation*}
\mathbf{x}^\ast \leftarrow clip(\mathbf{x} - \epsilon\cdot sgn(\nabla_{\mathbf{x}}  loss( -\mathbf{y}',H'(\mathbf{x} )) )),
\end{equation*}
where $\epsilon$ is the hyperparameter controlling the distortion under the $L_{\infty}$ norm. $clip(\mathbf{x})$ is utilized to clip each pixel of the images to the range of $[0, 255]$. $\mathbf{y}'$ is the vector defined in Equation~\eqref{equ:type1_2}, $H'$ is the classifier corresponding to the predictor $F'$ define in Equation~\eqref{equ:type1_2}. The $Loss$ is defined as sigmoid loss for each label.

\item \textbf{Targeted Fast Gradient Method (FG):} A variation of FGS using $L_2$ normalization as follows:
\begin{equation*}
\mathbf{x}^\ast \leftarrow clip(\mathbf{x} - \epsilon\cdot \frac{\nabla_{\mathbf{x}} loss( -\mathbf{y}' , H'(\mathbf{x} ))  }{ \| \nabla_{\mathbf{x}} loss( -\mathbf{y}' , H'(\mathbf{x} ) ) \|_2} ).
\end{equation*}

\item \textbf{ML-CW:} Our multi-label extension of Carlini \& Wagner multi-class targeted attack based on Equation~\eqref{equ:type1_3}. $L_2$ norm is applied.
\item  \textbf{ML-DP:} Our extension of DeepFool on multi-label adversarial attacks with Algorithm~\ref{alg:fool}.
\item   \textbf{Rank \uppercase\expandafter{\romannumeral1}:} The simplified ranking-based method based on Equation~\eqref{equ:rank2}. $L_2$ norm is applied to the distortion term.
\item   \textbf{Rank \uppercase\expandafter{\romannumeral2}:} A more sensitive ranking-based attack based on Equation~\eqref{equ:rank4}. $L_2$ norm is applied to the distortion term.
\end{itemize}

\renewcommand{\arraystretch}{1.2}
\begin{table*}[t!]\scriptsize
 \centering
  \caption{  General performance of all six methods with different attacking strategies in terms of three types of measurements.}
    \begin{tabular}{c|c||c|*{3}{c}|*{4}{c}|c} \toprule
 \multicolumn{2}{c||}{} & Distortion &\multicolumn{3}{c|}{Classification} &\multicolumn{4}{c|}{Ranking}  & \\ \cline{1-11}  
 \multicolumn{2}{c||}{Metrics} & RMSD $\downarrow$  & Hamming (A)  $\downarrow$ & Hamming (B)  $\downarrow$  & F1   $\uparrow$  & Ranking Loss $\downarrow$ & mAP $\uparrow$ & AUC $\uparrow$ &Kendall $\tau_b$   $\uparrow$   &   \multirow{13}{*}{\rotatebox[origin=c]{270}{VOC 2007}}         \\  \cline{1-10}  

  \multirow{6}{*}{ \shortstack{Random \\ Case,\\ $|A|=2$, \\$|B|=18$}} 
 		& FGS    & $ 5.9271 $   &$ 0.9045$  &$  0.0064 $  &$0.1420 $ 
		                         &$ 0.4113 $   &$ 0.2709 $    &$ 0.5887 $    & $ 0.0711 $    \\   
                                                     
                 & FG       & $5.8718 $   &$ 0.9100 $  &$  0.0183 $  &$ 0.1587  $ 
                                          &$ 0.4304 $   &$  0.2741  $    &$  0.5696 $    & $ 0.0670  $      \\                                
 		& ML-CW  & $ {\bf 2.4452}$     & $ 0.0500 $& ${\bf 0.0039}$ & $0.9253$ 
		                                & $0.0013$ & $0.9917$ & $0.9987$ &    $0.3571$ \\       

                 & ML-DP  & $ 11.0673$   &$ 0.7145 $  &$ 0.0084  $  &$0.1198$ 
                                                 &$ 0.4748 $   &$0.2243$    &$0.5252$    & $0.0269$      \\

		& Rank \uppercase\expandafter{\romannumeral1}    & $3.2108 $    &$ 0.1900 $  &$  0.0417 $  &$0.6610$ 
		                                &$ 0.0087  $   &$0.9308 $    &$0.9913$    & $0.3520$     \\     
                 & Rank \uppercase\expandafter{\romannumeral2}   & $ 5.2193$  &$ {\bf 0.0000} $  &$ 0.0056  $  &$ {\bf0.9700}$ 
                                                 &$ {\bf 0.0000} $   &${\bf 1.0000}$    &${\bf 1.0000}$    & ${\bf 0.3580}$       \\     \cline{1-10}

  \multirow{6}{*}{ \shortstack{Extreme\\ Case, \\$|A|=20$, \\$|B|=0$}}
 		& FGS    & $9.8397 $   &$ 0.9425  $  &$N.A.$  &$ 0.0779$
		                         &$ 0.9591 $   &$0.7799$    &$0.0409$    & $-0.4026$    \\                                                             
                 & FG       & $ 8.1945$  &$ 0.9440  $  &$ N.A.  $  &$0.0677$ 
                                          &$ 0.9529 $   &$ 0.7797  $    &$0.0471$    & $-0.3970$      \\                                
 		& ML-CW  & $ 9.1861 $       & ${\bf 0.5505} $    & $ N.A.$   & ${\bf 0.5419}$ 
		                                & $0.1413$   & $0.9816$      & $0.8587$      & $0.3149$ \\       
                 & ML-DP       & $ 19.2335 $   &$ 0.9160  $  &$ N.A.  $  &$ 0.0752 $ 
                                          &$ 0.8878 $   &$ 0.7971 $    &$ 0.1122 $    & $-0.3405$      \\    
		& Rank \uppercase\expandafter{\romannumeral1}    & $  {\bf 3.9147} $    &$ 0.8450 $  &$ N.A.  $  &$ 0.1069 $ 
		                                &$ 0.0692 $   &$0.9923$    &$0.9308$    & $0.3777$     \\     
                 & Rank \uppercase\expandafter{\romannumeral2}   & $6.2860 $    &$  0.8760$  &$ N.A.  $  &$ 0.0458 $ 
                                                 &$ {\bf 0.0675} $   &${\bf 0.9924} $    &${\bf 0.9325}$    & ${\bf 0.3793}$       \\   \bottomrule   \toprule                                   
                                                 


  \multirow{6}{*}{ \shortstack{Random \\ Case,\\ $|A|=2$, \\$|B|=18$}}
		 & FGS    & $5.9365  $    &$ 0.9400 $  &$  0.0117 $  &$ 0.1663  $ 
		                         &$ 0.4231 $   &$0.2827$    &$0.5769$    & $0.0725$  &   \multirow{12}{*}{\rotatebox[origin=c]{270}{VOC 2012}}  \\                              
		                                                                              
                 & FG       & $ 5.8904$    &$ 0.8900 $  &$ 0.0083  $  &$0.1374$ 
                                          &$ 0.4183 $   &$0.2653$    &$0.5817$    & $0.0662$      \\    
 		& ML-CW  & $  {\bf2.3928} $      & $ 0.0165$    & ${\bf 0.0008} $   & ${\bf 0.9746}$ 
		                                & $0.0003$   & $0.9972$      & $0.9997$      & $0.3443$ \\       

                    & ML-DP      & $10.2170 $    &$ 0.8750 $  &$  0.0194 $  &$0.1393$ 
                                          &$ 0.4866 $   &$0.2275$    &$0.5134$    & $0.0232$      \\                                                     
                                                  
		& Rank I    & $ 4.1550$    &$ 0.2010 $  &$ 0.0328  $  &$0.6180$ 
		                                &$ 0.0041 $   &$0.9789$    &$0.9959$    & $0.3416$     \\     
                 & Rank II   & $5.3902 $    &${\bf 0.0090} $  &$ 0.0020  $  &$0.9741$ 
                                                 &$ {\bf 0.0001} $   &${\bf 0.9993}$    &${\bf 0.9999}$    & ${\bf 0.3445}$       \\     \cline{1-10}

  \multirow{6}{*}{ \shortstack{Extreme\\ Case,\\$|A|=20$, \\$|B|=0$}}
 		& FGS    & $ 9.8387 $    &$ 0.9420 $  &$  N.A. $  &$0.0881$ 
		                         &$ 0.9427 $   &$ 0.8548 $    &$0.0573$    & $-0.3046$    \\                                                             
                 & FG       & $8.3135 $   &$  0.9460 $  &$  N.A. $  &$0.0765$ 
                                          &$ 0.9383 $   &$0.8562$    &$0.0617$    & $-0.3023$      \\                                
 		& ML-CW  & $ 9.9749 $      & $ {\bf 0.0520} $    & $ N. A. $   & ${\bf 0.9684}$ 
		                                & $0.0120$   & $0.9993$      & $0.9880 $      & $0.3383$ \\       
                 & ML-DP       & $ 20.5351 $    &$0.9120  $  &$ N.A. $  &$ 0.0943 $ 
                                          &$ 0.8648  $   &$0.8784$    &$0.1352$    & $-0.2477$      \\  
		& Rank \uppercase\expandafter{\romannumeral1}    & $ {\bf2.8764}$    &$ 0.8965 $  &$ N.A.  $  &$ 0.0806 $ 
		                                &$ 0.0071 $   &$ 0.9996 $    &$0.9929$    & $0.3419$     \\     
                 & Rank \uppercase\expandafter{\romannumeral2}   & $ 4.8311 $   &$ 0.9030 $  &$ N.A.  $  &$ 0.0675$ 
                                                 &$ {\bf 0.0066} $   &${\bf 0.9996}$    &${\bf 0.9934}$    & ${\bf 0.3422}$       \\   \bottomrule                 
    \end{tabular}
\label{table:Preform1}
 \vspace{-2pt}
\end{table*}

\noindent\textbf{Parameter Setting:}
Since binary search is applied for ML-CW, Rank \uppercase\expandafter{\romannumeral1}, and Rank \uppercase\expandafter{\romannumeral2} to find the suitable trade-off parameter $\lambda$, we set the initial $\lambda$ as $10^{5}$ and apply the binary search ten times. For each searched $\lambda$, the maximum iteration for finding the corresponding perturbation is set as $1000$. Learning rate is set as $10^{-2}$. To avoid the optimization getting stuck in extreme spots, we follow the similar image preprocessing and variable transformation methods discussed in~\cite{carlini2017towards}. For ML-DP, the maximum iteration is set as twenty to prevent over distortion. For FGS and FG, the hyperparameter $\epsilon$ is searched between $[0,10]$, and defined as the value which result the best attacking performance in different scenarios.





\subsection{Evaluation Metrics}

Three types of metrics are employed. 
 \textbf{Distortion:} root mean square deviation (RMSD)~\cite{liu2016delving} is used to measure the perturbation sizes. For each method, we report the RMSD, which could result in the best general performance of the classification and ranking tasks. \textbf{Classification:} we set the classification threshold as $0.5$ for each label, i.e., $h_i(\mathbf{x}^\ast) = 1$ if $f_i(\mathbf{x}^\ast)\ge0.5$. The classification performance is measured by Hamming Loss and instance-F1 score (F1)~\cite{wu2016unified}.  Hamming loss is applied on two targeted label sets $A$ and $B$. Since we focus on targeted attacks, the targeted labels are considered as the ground truth in classification attacks. 
\textbf{Ranking:} the ranking performance is measured by four metrics, i.e., Ranking Loss, mean average precision (mAP),  instance average area under curve (AUC), and Kendall $\tau_b$ rank correlation coefficient. For clarity, an arrow sign $\uparrow$ or $\downarrow$, is annotated behind each metric in tables indicating the higher the better ($\uparrow$), or the lower the better ($\downarrow$). It is worth pointing that we control the upper bound of the RMSD to search the best perturbations for all the methods in order to give a fair comparison on the classification and ranking performance.

\subsection{Experimental Settings}
\label{sec:exp_set}
To give a relatively comprehensive evaluation of six attacking methods, different combinations of attacking labels are chosen. For each dataset, we fix the classification threshold as $0.5$ and collect target attacking images into a set $X$. Every label of these images should be correctly classified by $H$. The attacking strategies employed in two parts of our experiments are introduced respectively as follows.


\noindent\textbf{General Attacking:}  we first try two types of attacking strategies to test the general performance of all methods:
\begin{itemize}
\itemsep-0.2em 
\item \textbf{Random Case:} Randomly select 1000 images from $X$. For each image $\mathbf{x}$, randomly select one positive and one negative label from sets $Y_1$ and $Y_{-1}$ defined in Table~\ref{table:div} as the targeted changing set $A$, and put the rest in $B$. 

\item \textbf{Extreme Case:}  Randomly select 1000 images from $X$. 
For each image, the goal is to change all labels, i.e., $A=L$ and $B=\emptyset$.
\end{itemize}

\noindent\textbf{Label Specified Attacking:}  In the second part of the experiment, two more strategies are used to evaluate six methods on attacking specific labels. To avoid lacking of test examples caused by label imbalance,  we choose the highest contained label ``person" as the targeted reduced label and augment the lowest contained label ``sheep":
\begin{itemize}
\itemsep-0.2em 
\item \textbf{Person Reduction (Person):} Randomly select 100 images from $X$. Each image should have at least two positive labels and one of them should be ``person". Set $A=\{person\}$ and $B=L\setminus\{person\}$.
\item \textbf{Sheep Augmentation (Sheep):} Randomly select 100 images from set $X$. Each of them should not have label ``sheep". Set $A=\{sheep\}$ and $B=L\setminus\{sheep\}$.
\end{itemize}

\subsection{General Attacking Performance}
We first compare the performance of all six methods with general attacking strategies. From results shown in Table~\ref{table:Preform1}, the main observations are described as follows.

\noindent\textbf{Distortion Analysis.}
We observe that: (1) ML-CW method provides the best RMSD in the ``Random Case'', but becomes worse than Rank I and Rank II in ``Extreme Case''. This is because the classification constraints used by ML-CW is easier to achieve than ranking-based constraints when the number of labels we need to change, i.e., $|A|$, is small, but will become much harder to achieve when $|A|$ is large. 
(2) The sizes of the perturbations generated by Rank I \& II are relatively stable in both cases. It is because the ranking-based attacks only care about the ranking relationships among labels without considering specific classification threshold. When $|A|$ is small, this might be harder to achieve compared with the classification constraints used by ML-CW and will cause the distortion to be larger. But when $|A|$ is large as in the ``Extreme Case'', the soft ranking thresholds would become milder than hard classification thresholds.  (3) ML-DP generates the largest perturbations in all cases because multiple linear constraints are hard to be jointly accommodated in multi-label settings.




\renewcommand{\arraystretch}{1.3}
\begin{table*}[t!]
 \centering
    \caption{Performance of the best three methods for label specified attacking on the VOC 2007 dataset.}
    \begin{tabular}{c|c||c|*{3}{c}|*{4}{c}} \toprule
 \multicolumn{2}{c||}{} & Distortion &\multicolumn{3}{c|}{Classification} &\multicolumn{4}{c}{Ranking}   \\  \cline{1-10}  
 \multicolumn{2}{c||}{Metrics} & RMSD $\downarrow$  & Hamming (A)  $\downarrow$ & Hamming (B)  $\downarrow$  & F1   $\uparrow$  & Ranking Loss $\downarrow$ & mAP $\uparrow$ & AUC $\uparrow$ &Kendall $\tau_b$   $\uparrow$  \\  \cline{1-10}  

    \multirow{3}{*}{ \shortstack{Person,\\ $|A|=1$, \\$|B|=19$}}
 		& ML-CW  & $ 1.5735$        & $ 0.0700 $    & $ 0.0032 $   & $0.9300$           
		                                & $ 0.0005 $   & $0.9950$      & $0.9995$      & $0.3195$ \\       
		& Rank \uppercase\expandafter{\romannumeral1} &${\bf 0.1382} $    &$  0.9800 $  &$  0.0116$  &$ 0.6432  $   
		                                &$  0.0021$   &$0.9800$    &$0.9979$    & $0.3185$     \\     
                 & Rank \uppercase\expandafter{\romannumeral2}   & $ 5.2107 $  &$ {\bf 0.0000}   $  &$  {\bf 0.0000} $  &${\bf 1.0000}$   
                                                 &$ {\bf 0.0000} $   &${\bf 1.0000}$    &${\bf 1.0000}$    & ${\bf 0.3198}$       \\     \hline                                                 
  \multirow{3}{*}{ \shortstack{Sheep,\\ $|A|=1$, \\$|B|=19$}}
 		& ML-CW  & $ {\bf 1.9293} $     & $ {\bf 0.0800} $    & ${\bf 0.0032} $   & ${\bf 0.9690} $    
		                                &$ 0.0008 $   &$0.9942$    &$0.9992$    & $0.4613$       \\        
		& Rank \uppercase\expandafter{\romannumeral1}    & $3.0393 $  &$0.8300  $  &$0.0042   $  &$0.7360$   
		                                &$  0.0037$   &$0.9817$    &$0.9963$    & $0.4586$     \\     
                 & Rank \uppercase\expandafter{\romannumeral2}   & $ 3.4916 $ &$ 0.1500 $  &$ 0.0184  $  &$0.8112$   
                                                 & $ {\bf 0.0000} $   & ${\bf 1.0000}$      & ${\bf1.0000}$      & ${\bf 0.4620}$ \\  \bottomrule    

    \end{tabular}
\label{table:Perform2}
 \vspace{-5pt}
\end{table*}


\noindent\textbf{Attack Classification.}
From the results measured by three classification metrics, we can see that: (1) Two one-step methods FG and FGS do not perform well. This shows that one-shot methods are not good at generating multi-label adversarial examples. (2) In general, ML-CW performs the best among all six methods. Rank II performs comparably well in the ``Random Case'', but becomes worse in the ``Extreme Case''. This is because the ranking-based methods do not consider the hard classification threshold $0.5$ during the optimization while ML-CW does on the contrary. It should be noted that this does not mean Rank II cannot perform well in this case. As mentioned in Section~\ref{sec:rank_general}, Rank II could also take the classification thresholds into account by setting fixed thresholds in the constraints. 
(3) Rank \uppercase\expandafter{\romannumeral2} outperforms Rank \uppercase\expandafter{\romannumeral 1} in the ``Random Case'', but becomes slightly worse in the ``Extreme Case''.  By comparing their loss functions, we know that Rank \uppercase\expandafter{\romannumeral2} has more constraints in discriminating the labels within sets $\Omega^{-1}$ and $\Omega^{+}$. However, more ranking constraints bring more soft thresholds, which make it worse in the ``Extreme Case''. 
%



\noindent\textbf{Attack Ranking.} Three main observations can be summarized from Table~\ref{table:Preform1}. (1) FG and FGS cannot provide satisfying attacks due to their poor capacities in accommodating multiple labels. (2) Rank \uppercase\expandafter{\romannumeral2} outperforms all other methods. Comparing with Rank \uppercase\expandafter{\romannumeral1}, it enjoys more hierarchies among labels which could be reflected by the Kendall $\tau_b$ metric. (3) ML-CW performs well in general. However, it requires larger distortion to achieve comparable performance with ranking-based approaches in the ``Extreme Case''.  

\subsection{Label Specified Attacking Performance}
\begin{figure}[t!]
\vspace{5pt}
\hspace{0.15em}
\includegraphics[width=8.4cm]{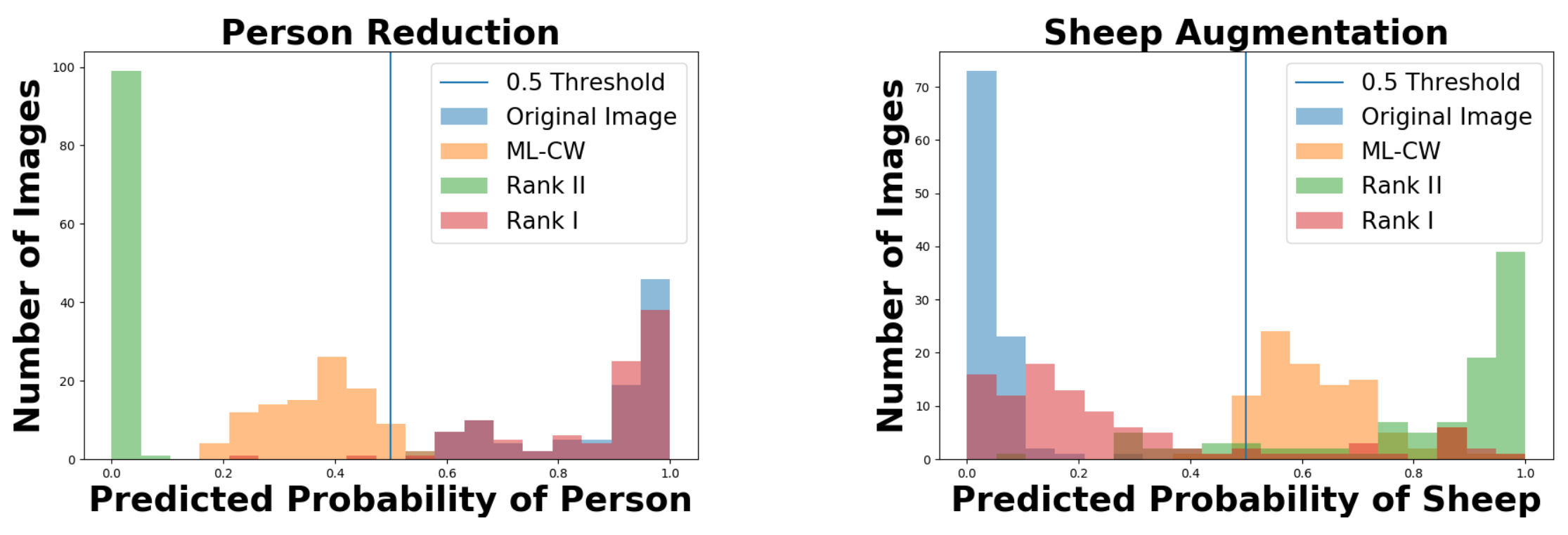}\\
\vspace{-10pt}
\caption{Performance on attacking person and sheep labels.}
\vspace{5pt}
\label{fig:ps}
\vspace{-16pt}
\end{figure}

We now adopt two more specific attacking strategies, which has been introduced Section~\ref{sec:exp_set}, i.e., ``Person Reduction" and ``Sheep Augmentation", to provide a more exquisite evaluation. We focus on the best three methods ML-CW, Rank I, and Rank II and describe the results in in Table~\ref{table:Perform2}. To give a more visualized comparison, we also display four probability distributions of the two target labels of all adversarial images generated by different methods in Fig.~\ref{fig:ps}. The X-axis represents the predictions of ``person'' or ``sheep''. The Y-axis denotes the number of images. Based on Table~\ref{table:Perform2} and Fig.~\ref{fig:ps}, we analyze the three methods in turn.




\noindent\textbf{ML-CW.} Two major observations are found. (1) ML-CW performs well in both cases since the number of labels we want to vary (i.e., $|A|$) is small. (2) As shown in Fig.~\ref{fig:ps}, ML-CW successfully decreases the ``person'' predictions of most images below the classification threshold $0.5$, and increases the predictions of ``sheep'' beyond $0.5$. However, the mean values of the distributions in both cases are close to $0.5$ because ML-CW only considers the linear error between predictions and the thresholds, which lacks discriminative power.




\noindent\textbf{Rank I Attack.} From Table~\ref{table:Perform2},  we find that the perturbations generated by Rank I are pretty small in ``person'' case.  This validates our discussion in Section~\ref{sec:rank2}, i.e., as long as ``person'' ranks lower than other originally positive labels and higher than the rests, the optimization will stop,  which causes the distortion being small but the probability of ``person'' remaining high. This explanation can be further verified by Fig.~\ref{fig:ps}. For both cases, the ``person'' predictions of the adversarial images generated by Rank I (red bars) have high overlap with the predictions of the original images (blue bars).

\noindent\textbf{Rank II Attack.} Two main conclusions could be drawn as follows. (1) In Fig.~\ref{fig:ps}, Rank II not only successfully decreases the predictions of ``person'' below the classification threshold, but also induces a huge discrimination between the mean value and the classification threshold. Similar results happen in the ``sheep'' case. (2) For the classification attack, Rank \uppercase\expandafter{\romannumeral2} performs better in eliminating ``person'' than augmenting ``sheep''. The reason lies in that every image in the dataset has much more negative labels than positive ones. Due to the lack of positive labels, during the optimization process in ``sheep'' attack, it is hard to guarantee that there will always be some labels whose predicted probabilities are larger than $0.5$. Thus, even ``sheep'' ranks the highest among all labels, its prediction may still be smaller than $0.5$ with high probability. 

\subsection{Discussion}
In the section, we introduce two interesting findings, which could potentially benefit the future in-depth analysis of the interpretability and robustness in multi-label learning.


\subsubsection{Case Study}
\begin{figure}[t!]
\vspace{0pt}
\begin{minipage}[b]{0.5\textwidth}
\includegraphics[width=8.5cm, height=3.2cm]{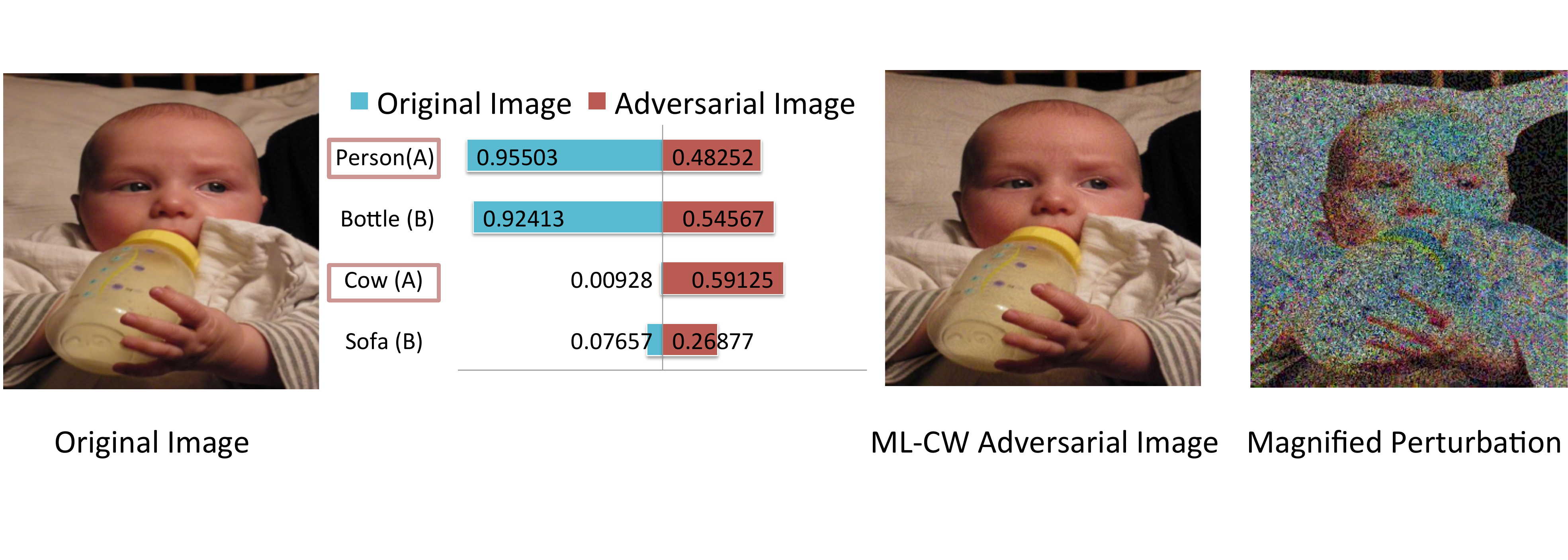}\\
\vspace{-25pt}
\end{minipage}%

\begin{minipage}[b]{0.5\textwidth}
\includegraphics[width=8.5cm,height=3.2cm]{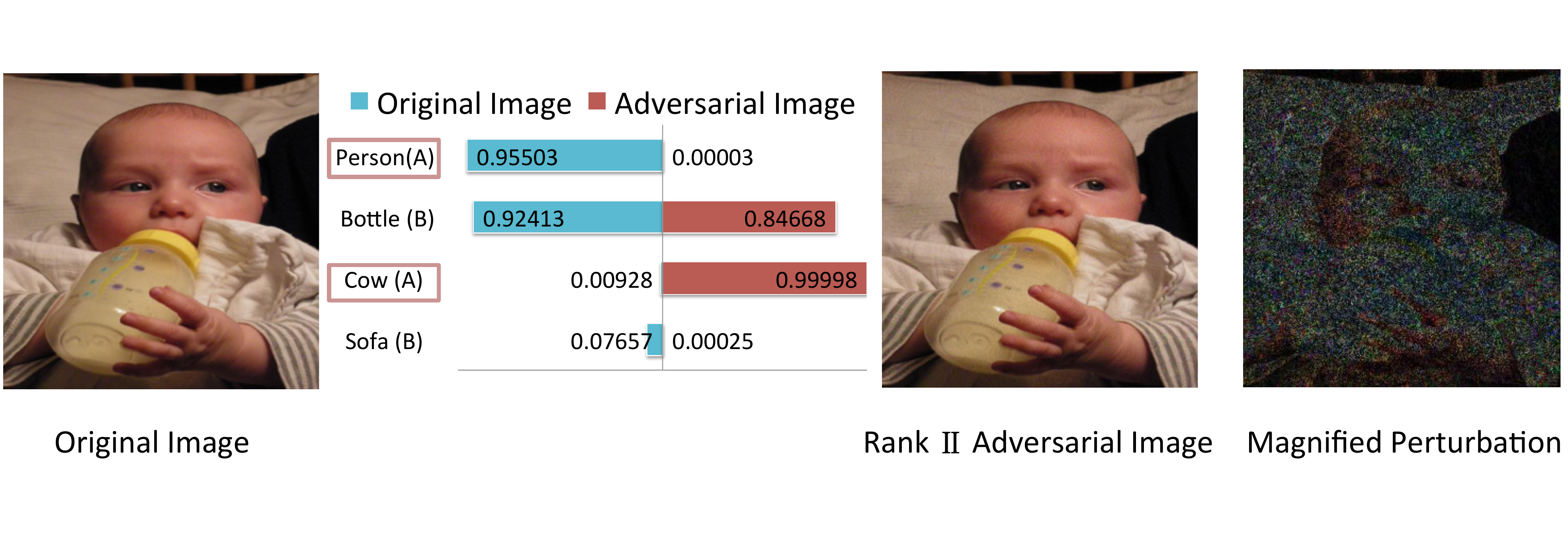}\\
\end{minipage}%
\vspace{-23pt}
\caption{Comparison of the attacking performance and generated perturbations of ML-CW and Rank II.}
\label{fig:case}
\vspace{-5pt}
\end{figure}

Fig.~\ref{fig:case} compares the attacking performance of ML-CW and Rank II, and their generated perturbations. We discuss the findings one by one corresponding to the four parts of the figure from left to right. First, from the leftmost part, we can see that the original image has two labels, i.e., ``person'' and ``bottle''. The attacking strategy is to degrade ``person'' and highlight ``cow'' while fixing ``bottle'' and ``sofa'', i.e., $A=\{person, bottle\}$ and $B=\{cow, sofa\}$. The remaining labels are defined as set $C$.
Second, the histograms in the second part compare the predictions of four labels before and after the attacks. If the classification threshold is $0.5$, both methods would successfully achieve the purpose. However, Rank II gives higher discrimination (variance) among the four labels. Third, if we compare the original image with two adversarial images, the changes are pretty small and human imperceptible. It provides the evidence that the multi-label model we attack is vulnerable and not robust. Fourth, to give an intuitive visualization of the generated perturbations, we display the magnified perturbations generated by each method in the rightmost part. Since the perturbations are too small to be observed, we first get the absolute value of each pixel  $x_i$ and then magnify them by multiplying a constant $255.0/\max\{x_i| i =1,\ldots,d \}$. As we can see, the perturbations give a high-level reflection of the original image. This shows the generative power of adversarial examples leveraging the adversarial relationship between the attacking model and its targeted model. Also, ML-CW gives a clearer reflection than Rank II, which is probably caused by the hard classification threshold we induced.




\subsubsection{A Simple Defense Strategy}

\begin{figure}[t!]
\begin{minipage}[b]{0.25\textwidth}
\includegraphics[width=4.5cm]{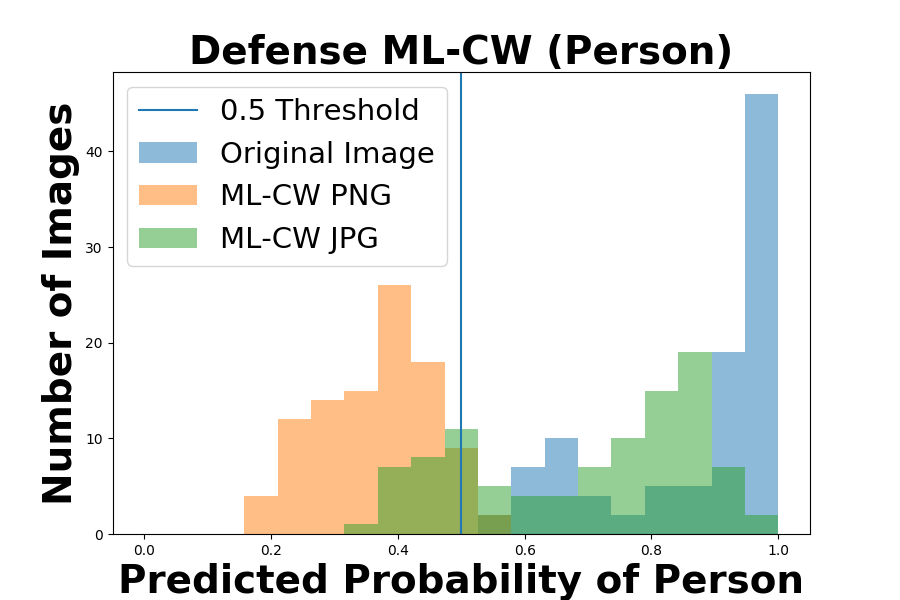}\\
\end{minipage}%
\begin{minipage}[b]{0.25\textwidth}
\includegraphics[width=4.5cm]{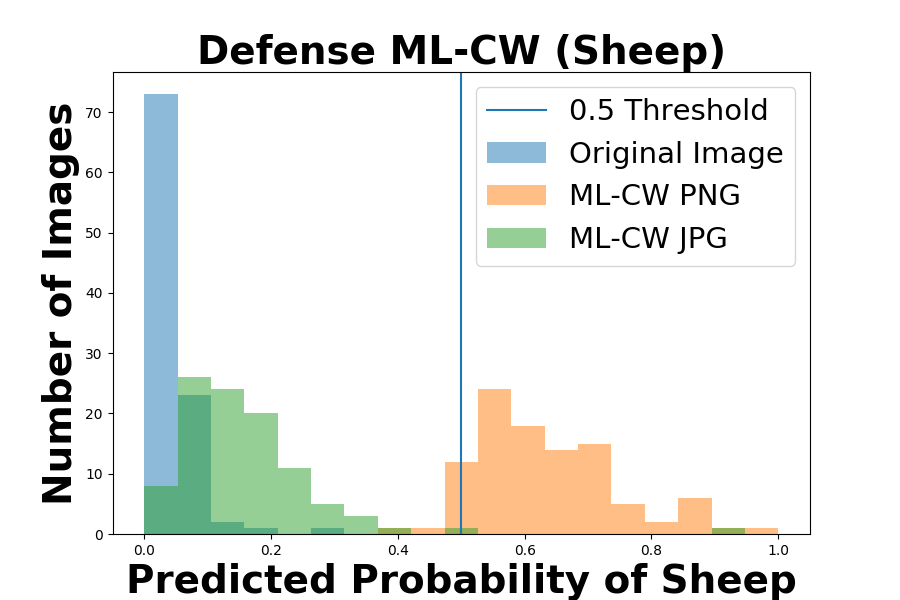}\\
\end{minipage}%
\\[0cm]
\begin{minipage}[b]{0.25\textwidth}
\includegraphics[width=4.5cm]{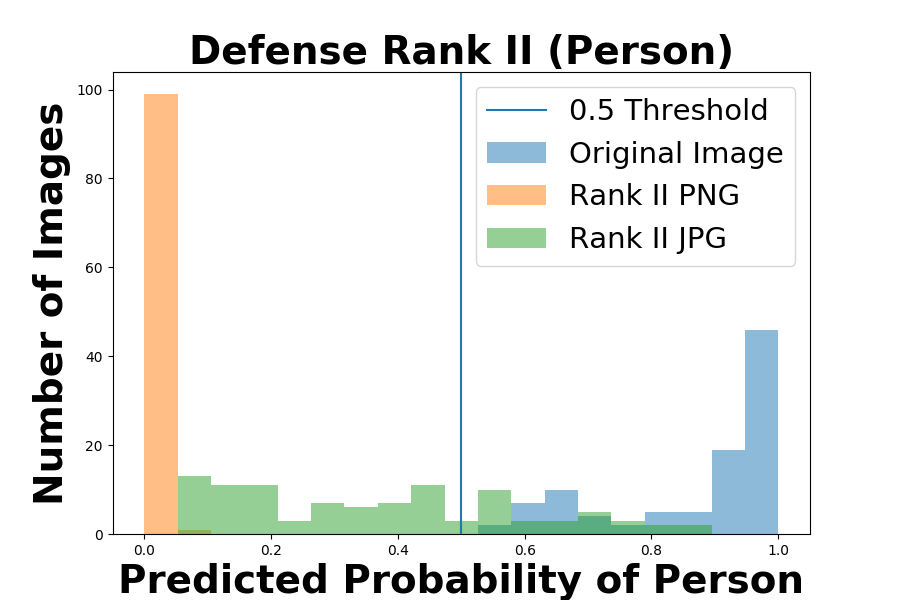}\\
\end{minipage}%
\begin{minipage}[b]{0.25\textwidth}
\includegraphics[width=4.5cm]{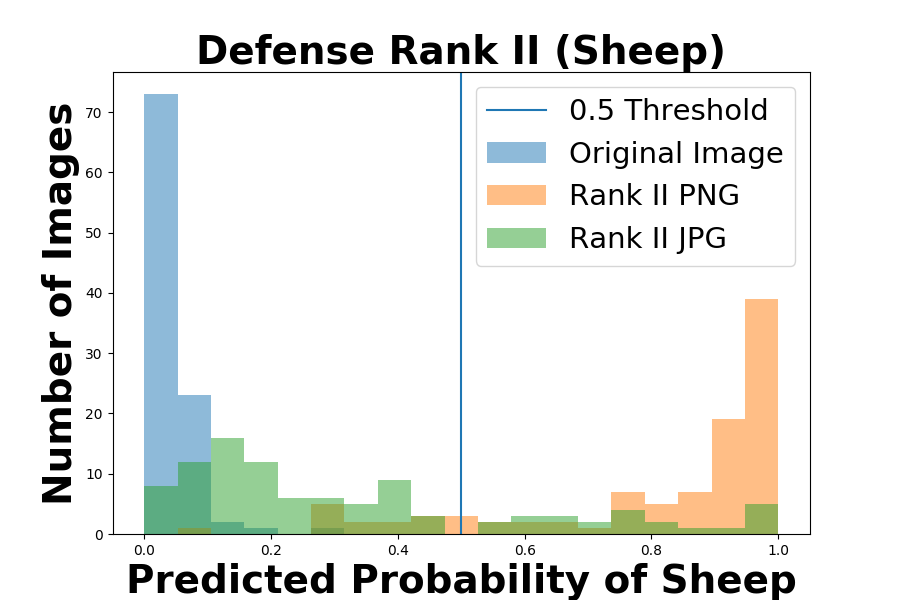}\\
\end{minipage}%
\vspace{-10pt}
\caption{JPEG decoding defense on multi-label attacks.}
\label{fig:defense}
\vspace{-5pt}
\end{figure}

To mitigate the threats of adversarial examples, various defense attempts have been made such as adversarial training~\cite{goodfellow2015explaining}, defensive distillation~\cite{papernot2016distillation} and some detection-based methods~\cite{metzen2017detecting}. One practical way, which could be used in combination with any machine learning models, is to pre-process input examples to purify the possible perturbations. During our experiments, an interesting observation attracts us is that: JPEG encoding could defense the diverse multi-label adversarial attacks to some extent. Fig.~\ref{fig:defense} shows the prediction histograms of the PNG adversarial images generated by ML-CW and Rank II, and their JPEG compression on the label specified attacks. We can see that for both attacking methods and labels, the JPEG histogram has higher overlapped areas with the histogram of original images than PNG images. It illustrates that the effect of multi-label perturbations is largely reduced based on the JPEG decoding. Though this defense strategy is not always success especially for larger perturbations, it is still a practical solution since it requires no extra detecting and training procedure or specification on targeted labels. Related researches can also be found in~\cite{dziugaite2016study,kurakin2017adversarial,das2017keeping}.

 \section{Related Work}
The related work can be categorized into two main topics: \textbf{Adversarial Attacks} and \textbf{Multi-label Learning}.
\subsection{Adversarial Attacks}  
 The exploration of adversarial attacks roots in the demand of understanding the vulnerability and robustness of machine learning models~\cite{barreno2006can}. By exploiting the adversarial games between machine learning models and attackers~\cite{papernot2016towards}, effective attacks could be made to sabotage machine learning models. Though adversarial attacking has been studied for a long time, the term of adversarial example is first coined by Szegedy et al.~\cite{szegedy2014intriguing} to refer to the perturbed inputs which mislead a classifier. Subsequently, vast approaches have been proposed, which are mainly gradient-based methods focusing on attacking deep learning models. From the optimization perspective, these methods branch into two categories, i.e., either the one-shot methods such as the FGS method~\cite{goodfellow2015explaining} or the iterative optimization approaches such as the L-BFGS method~\cite{szegedy2014intriguing}, iterative least likely method~\cite{kurakin2017adversarial}, Jacobian-based saliency map attack~\cite{papernot2016limitations}, and so on~\cite{liu2016delving,huang2017adversarial,biggio2013evasion}. From the application perspective, most of them focus on image-related data~\cite{carlini2017towards,moosavi2016deepfool}, while some recent work also investigates other data types such as text data~\cite{jia2017adversarial}. Though these methods are mainly proposed for white-box attack as what we target on in this paper, they can also be used in black-box settings by leveraging the transferability among models~\cite{liu2016delving,papernot2017practical}. Besides, adversarial attacks are also applied in the training process to enhance the robustness of various models~\cite{goodfellow2015explaining,kurakin2016adversarial,wu2017adversarial} and utilized towards the interpretability of deep neural networks~\cite{dong2017towards,tabacof2016exploring,tramer2017space}. Though various attempts have been on proposing either un-targeted or targeted attacks, to our best knowledge, most of them still rests on the multi-class setting which could not be directly applied to the multi-label cases.

\subsection{Multi-label Learning.}
Generalizing from multi-class setting, multi-label learning has become a popular learning paradigm in recent years due to its abundant applications in extensive areas~\cite{gibaja2015tutorial,thabtah2004mmac, read2008multi}. Two main tasks in multi-label learning are multi-label classification and multi-label ranking. Multi-label classification aims to assign the relevant labels to each instance form a specified label set~\cite{tsoumakas2006multi}. Problem transformation methods~\cite{ tsoumakas2006multi} are the most intuitive category of methods which aims to transform the multi-label classification into multiple binary classification problems (e.g., binary relevance approach) or a single multi-class classification problem (e.g., label powerset approach).  Classifier chains~\cite{read2011classifier} is also an alternative ensembling approach belonging to this category, which emphasizes the relationships among labels. Another popular category of methods is the algorithm adaptation methods, which adapts the algorithms to directly perform multi-label classification~\cite{zhang2007ml,chen2003constructing,zhang2006multilabel}. Comparing with multi-label classification, multi-label ranking is a more general task aiming to generate a consistent ranking among labels and could be transferred into the multi-label classification grounded on bipartition thresholds~\cite{gibaja2015tutorial}. Besides traditional ranking-based approaches~\cite{tsoumakas2009mining, zhang2014review}, recent advances have also combined multi-label learning with adversarial learning and demonstrate the effectiveness multi-label adversarial examples for adversarial training~\cite{wu2017adversarial}.

\section{Conclusion and Future Work}
\label{Conclusion}

In this paper, we focus on a new problem of generating multi-label adversarial examples and propose two general frameworks targeting on attacking multi-label classification and ranking models, respectively. For each type of the framework, we propose two specific iterative methods to generate targeted multi-label adversarial examples. By conducting experiments with different attacking strategies on deep neural networks, we empirically validate the effectiveness of our proposed attacking methods and the vulnerability of multi-label deep learning models on various targeted attacks. The future work will lie in proposing effective defensive methods to improve the robustness of multi-label learning models. Further theoretical analysis on different attacking perturbations may enhance interpretability and security of multi-label learning models, which is also intriguing for future exploration.



\balance
\bibliographystyle{IEEEtran}
\bibliography{adv}

\end{document}